\documentclass[a4paper,11pt]{article}
\usepackage{graphicx}
\usepackage{authblk}
\usepackage{array}
\usepackage{varwidth}
\usepackage{graphicx}
\usepackage{amssymb}
\usepackage{algorithmic,algorithm}
\usepackage{bm}
\usepackage{amsthm}
\usepackage[dvipsnames]{xcolor}
\usepackage{amsmath} 			
\usepackage{url}
\usepackage{rotating}
\usepackage{lineno,hyperref}
\usepackage[font=small,labelfont=bf]{caption}

\DeclareMathOperator{\argmin}{arg~min}

\def\FfigSensors{\centering\includegraphics[width=0.9\columnwidth]{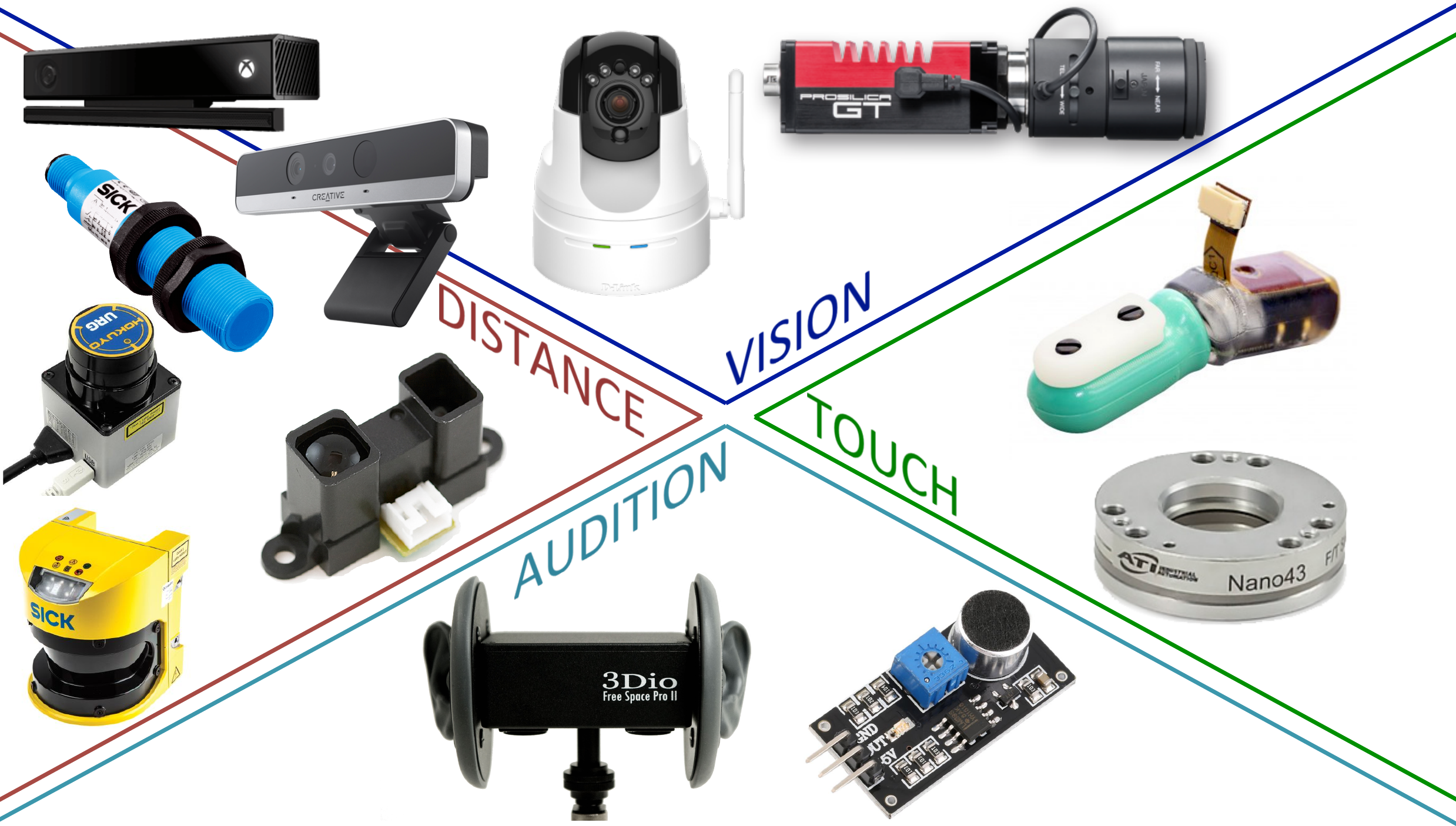}}
\def\FallServos{\centering\includegraphics[width=0.8\columnwidth]{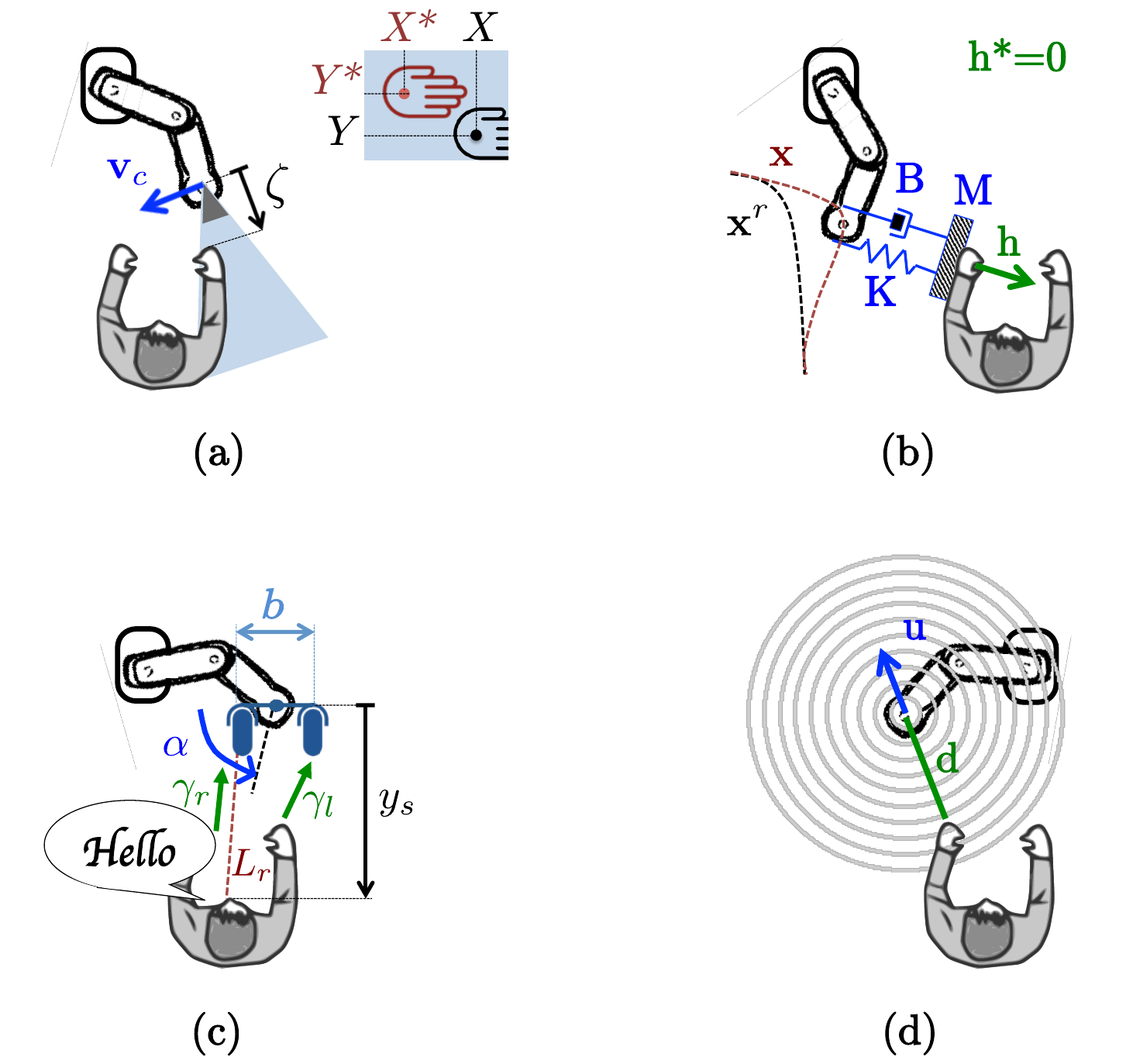}}

\def\FsharedVisionForce{\centering\includegraphics[width=1.0\columnwidth]{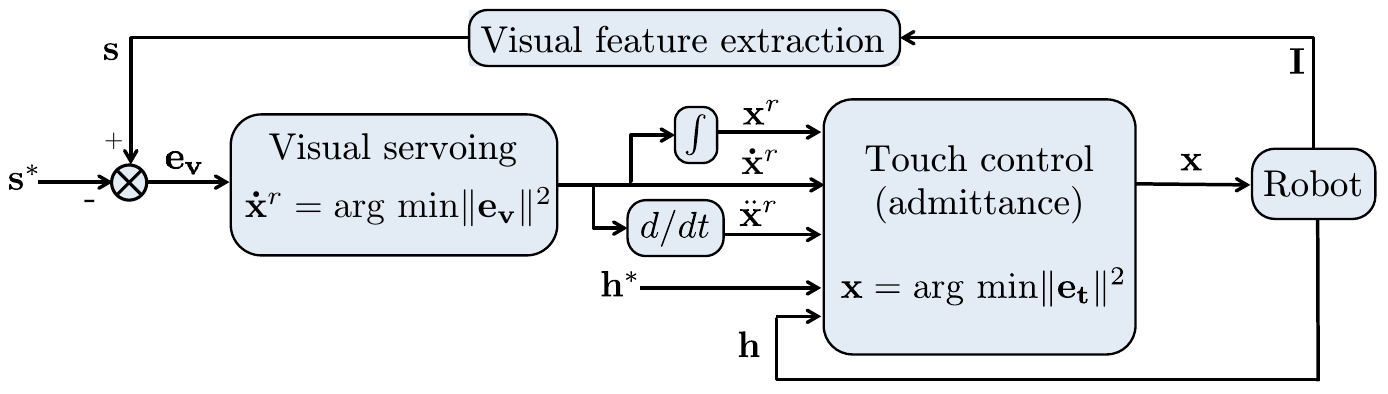}}
\begin{document}

\title{Sensor-Based Control for Collaborative Robots: Fundamentals, Challenges and Opportunities}
\author[a]{Andrea Cherubini}
\author[b]{David Navarro-Alarcon}
\affil[a]{Universit\'{e} de Montpellier / LIRMM, France}
\affil[b]{The Hong Kong Polytechnic University, Hong Kong}
\date{}

\maketitle

\begin{abstract}
The objective of this paper is to present a systematic review of existing sensor-based control methodologies for applications that involve direct interaction between humans and robots, in the form of either physical collaboration or safe coexistence.
To this end, we first introduce the basic formulation of the sensor-servo problem, then present the most common approaches: vision-based, touch-based, audio-based, and distance-based control.
Afterwards, we discuss and formalize the methods that integrate heterogeneous sensors at the control level.
The surveyed body of literature is classified according to the type of sensor, to the way multiple measurements are combined, and to the target objectives and applications.
Finally, we discuss open problems, potential applications, and future research directions.
\end{abstract}


\section{Introduction}
Robot control is a mature field: one that is already being heavily commercialized in industry.
However, the methods required to regulate interaction and collaboration between humans and robots have not been fully established yet. These issues are the subject of research in the fields of physical human-robot interaction (pHRI) \cite{BiPeCo:08} and collaborative robotics (CoBots) \cite{Colgate_Cobots}. The authors of \cite{LuFl:12} present a paradigm that specifies the three nested layers of consistent behaviors that the robot must follow to achieve safe pHRI: 
\begin{equation}
\left\{\left\{\left\{\textrm{safety}\right\},\textrm{coexistence}\right\},\textrm{collaboration}\right\} \nonumber
\end{equation}
\begin{itemize}
	\item \textit{Safety} is the first and most important feature in collaborative robots. Although there has been a recent push towards standardization of robot safety (e.g., the ISO~13482:2014 for robots and robotic devices~\cite{iso:robots}), we are still in the initial stages. Safety is generally addressed through \textit{collision avoidance} (with both humans or obstacles~\cite{Kh:85}), a feature that requires high reactivity (high bandwidth) and robustness at both the perception and control layers.
	\item \textit{Coexistence} is the robot capability of sharing the workspace with humans. This includes applications involving a passive human (e.g., medical operations where the robot is intervening on the patients' body~\cite{Azizian2014}), as well as scenarios where robot and human work together on the same task, without contact or coordination. 
	\item \textit{Collaboration} is the capability of performing robot tasks with direct human interaction and coordination. There are two modes for this: \textit{physical} collaboration (with explicit and intentional contact between human and robot), and \textit{contactless} collaboration (where the actions are guided by an exchange of information, e.g., in the form of body gestures, voice commands, or other modalities). 
	Especially for the second mode, it is crucial to establish means for intuitive control by the human operators, which may be non-expert users. 
	The robot should be proactive in realizing the requested tasks, and it should be capable of inferring the user's intentions, to interact more naturally from the human viewpoint.
\end{itemize}
All three layers are hampered by the unpredictability of human actions, which vary according to situations and individuals, complicating modeling \cite{Phoha:2014}, and use of classic control.

In the robotics literature, two major approaches for task execution have emerged: \textit{path/motion planning}~\cite{La:06} and \textit{sensor-based control}~\cite{chaumette:ram:2006}. The \textit{planning} methods rely on a priori knowledge of the future robot and environment states over a time window. Although they have proved their efficiency in well-structured applications, these methods are hardly applicable to human-robot collaboration, because of the unpredictable and dynamic nature of humans. It is in the authors' view that \textit{sensor-based control} is more efficient and flexible for pHRI, since it closes the perception-to-action loop at a lower level than path/motion planning. Note also that sensor-based control strategies strongly resemble the processes of our central nervous system~\cite{Be:02}, and can trace their origins back to the servomechanism problem \cite{Davison:1975}.
The most known example is image-based visual servoing~\cite{chaumette:ram:2006} which relies directly on visual feedback to control robot motion, without requiring a cognitive layer nor a precise model of the environment.   

The aim of this article is to survey the current state of art in \textit{sensor-based control, as a means to facilitate the interaction between robots, humans, and surrounding environments}. Although we acknowledge the need for other techniques within a complete human-robot collaboration framework (e.g., path planning as mentioned, machine learning, etc.), here we review and classify the works which exploit sensory feedback to directly command the robot motion.

The timing and relevance of this survey is twofold. On one hand, while there have been previous reviews on topics such as (general) human-robot collaboration~\cite{Ajoudani2017,Villani2018} and human-robot safety~\cite{AlDeHa:17}, there is no specific review on the use of \emph{sensor-based control} for human-robot collaborative tasks. 
On the other hand, we introduce a \emph{unifying} paradigm for designing controllers with four sensing modalities. This feature gives our survey a valuable tutorial-like nature.

The rest of this manuscript is organized as follows: Section 2 presents the basic formulation of the sensor-based control problem; Section 3 describes the common approaches that integrate multiple sensors at the control level. Section 4 provides several classifications of the reviewed works. Section 5 presents insights and discusses open problems and areas of opportunity. Conclusions are given in Section 6.

\begin{figure}[t!]
	\centering \FfigSensors
	\caption{Examples of artificial sensors. Clockwise from the top left: Microsoft Kinect$^{\textregistered}$ and Intel Realsense$^{\textregistered}$ (vision and distance), Sony D-Link DCS-5222L$^{\textregistered}$ and AVT GT$^{\textregistered}$ (vision), Syntouch BioTac$^{\textregistered}$ and ATI Nano 43$^{\textregistered}$ (touch), sound sensor LM393$^{\textregistered}$ and 3Dio Free Space Pro II$^{\textregistered}$ Binaural Microphone (audition), proximity sensor Sharp GP2Y0A02YK0F$^{\textregistered}$, Laser SICK$^{\textregistered}$, Hokuyo URG$^{\textregistered}$ and proximity sensor SICK CM18-08BPP-KC1$^{\textregistered}$ (distance). Note that Intel Realsense$^{\textregistered}$ and Microsoft Kinect$^{\textregistered}$ provide both the senses of vision and of distance.}
	\label{Fig:FfigSensors}
\end{figure}

\section{Sensing Modalities for Control}
Recent developments on bio-inspired measurement technologies have made sensors affordable and lightweight, easing their use on robots. 
These sensors include RGB-D cameras, tactile skins, force/moment transducers, etcetera (see Fig.~\ref{Fig:FfigSensors}). The works reviewed here rely on different combinations of sensing modalities, depending on the task at stake. We consider the following four robot senses:

\begin{itemize}
\item \textit{Vision}. This includes methods for processing and understanding images, to produce numeric or symbolic information reproducing human sight. Although image processing is complex and computationally expensive, the richness of this sense is unique. Robotic vision is fundamental for understanding the environment and human intention, so as to react accordingly.
	
\item \textit{Touch}. In this review, touch includes both \textit{proprioceptive force} and \textit{tact}, with the latter involving \textit{direct} physical contact with an external object. \textit{Proprioceptive force} is analogous to the sense of muscle force~\cite{GaPr:12}. The robot can measure it either from the joint position errors or via torque sensors embedded in the joints; it can then use both methods to infer and adapt to human intentions, by relying on force control~\cite{Ho:85,villani:hbr:2008,RaCr:81,morel:icra:1998}. Human \textit{tact} (somatosensation), on the other hand, results from activation of neural receptors, mostly in the skin. These have inspired the design of artificial tactile skins~\cite{LoJoSa:08,CaNaMa:11}, thoroughly used for human-robot collaboration. 
   
\item \textit{Audition}. In humans, localization of sound is performed by using binaural audition (i.e., two ears). They exploit auditory cues in the form of level/time/phase differences between left and right ears to determine the source's horizontal position and its elevation \cite{Rayleigh1907}. Microphones artificially emulate this sense, and allow robots to ``blindly'' locate sound sources. Although robotic hearing typically uses two microphones mounted on a motorized head, other non-biological configurations exist, e.g. a head instrumented with a single microphone or an array of several omni-directional microphones~\cite{Nakadai:2006}.

\item \textit{Distance}. 
This is the only sense among the four that humans cannot directly measure. Yet, numerous examples exist in the mammal kingdom (e.g., bats and whales), in the form of echolocation. Robots measure distance with \textit{optical} (e.g., infrared 
or lidar), \textit{ultrasonic}, or \textit{capacitive}~\cite{BlGoWo:10} sensors. 
The relevance of this particular ``sense'' in human-robot collaboration is motivated by the direct relationship existing between the distance from obstacles (here, the human) and safety. 
\end{itemize}

Roboticists have designed other bio-inspired sensors, to \textit{smell} 
(see~\cite{Kowadlo2008} for a comprehensive survey and~\cite{Russell2006,Gao2016,Rahbar2017} for 3D tracking applications) 
and \textit{taste}~\cite{Shimazu2007,Kobayashi2010,Ha2015}. 
However, in our opinion, artificial \textit{smell} and \textit{taste} are not yet mature enough for human-robot collaboration. Most of the current work on these senses is for localization/identification of hazardous gases/substances.
For these reasons, this review will focus only on the four senses mentioned above, namely vision, touch, audition and distance.


\section{Sensor-Based Control}
\label{chap:sensor-based_control}

\subsection{Bio-Inspired Strategy}
Currently, one of the most disruptive theories in cognitive science --- the  \textit{embodied cognition theory} --- states that sophisticated bodily behaviors (e.g. choreographed limb motions) result from perceptually-guided actions of the body and of its interaction with the environment \cite{Wilson:2013}. This revolutionary idea (referred by Shapiro as the ``replacement hypothesis'' \cite{Shapiro:2010}) challenges traditional cognitive science, by proposing the total replacement of complex symbolic task representations with simpler perception-to-action regulatory models.
A classic example of this theory is the baseball outfielder problem. For traditional cognitive science, it is solved by computing a physics-based simulation of the flying ball to predict its trajectory and landing point \cite{Saxberg:1987}. 
Instead, the embodied cognition counterpart formulates the solution as the explicit control of the optical ball movements (as perceived by the outfielder) by running lateral paths that maintain a linear optical trajectory, to anticipate the ball trajectory and perform a successful catch \cite{McBeath:1995}.
From the perspective of robotics, the latter approach is interesting as it clearly models the --- apparently complex --- motion task as a simple feedback control problem, which can be solved with sensor-based strategies, and without simulations or symbolic task representations. The following section reviews the basic formulation of sensor-based feedback control, since it is the most used in the papers that we reviewed. 

\subsection{Formulation of Sensor-Based Control}
Sensor-based control aims at deriving the robot control input $\mathbf{u}$ (operational space velocity, joint velocity, displacement, etc.) that minimizes a \emph{trajectory} error $\mathbf{e} = \mathbf e(\mathbf u)$, which can be estimated by sensors and depends on $\mathbf u$.
A general way of formulating this controller (accounting for actuation redundancy $\dim(\mathbf u) > \dim(\mathbf e)$, sensing redundancy $\dim(\mathbf u) < \dim(\mathbf e)$, and task constraints) is as the quadratic minimization problem:
\begin{equation}
\begin{aligned}
&\mathbf{u} = \underset{\mathbf{u}}{\argmin}~ \frac{1}{2}\|\mathbf e(\mathbf u)\|^2 \label{qp} \\
&\text{subject to task constraints.}
\end{aligned}
\end{equation}

This formulation encompasses the classic \textit{inverse kinematics} problem \cite{Whitney1969} of controlling the robot joint velocities ($\mathbf{u} = \dot{\mathbf{q}}$), so that the end-effector operational space position $\mathbf{x}$ converges to a desired value $\mathbf{x}^*$.
By defining the desired end-effector rate as $\dot{\mathbf x}^* = - \lambda \left( \mathbf x - \mathbf x^* \right)$, for $\lambda>0$, and setting $\mathbf{e} = \mathbf{J} \dot{\mathbf{q}} - \dot{\mathbf x}^*$ for $\mathbf J = \partial\mathbf x/\partial\mathbf q$ as the Jacobian matrix, it is easy to show that the solution to \eqref{qp} (in the absence of constraints) is $\dot{\mathbf q} = \mathbf J^+\dot{\mathbf x}^*$, with $\mathbf{J}^+$ the generalized inverse of $\mathbf J$. 
This leads to the set-point controller\footnote{Throughout the paper, $\lambda$ is a positive tuning scalar that determines the convergence rate of task error $\mathbf e$ to $\mathbf 0$.}:
\begin{equation}
\dot{\mathbf{q}} = - \mathbf J^+ \lambda \left(\mathbf x - \mathbf x^* \right).
\label{Eq:taskControl}
\end{equation}

In the following sections, we show how each of the four senses (\textit{vision}, \textit{touch}, \textit{audition} and \textit{distance}) has been used for robot control, either with~(\ref{qp}), or with similar techniques. Figure \ref{Fig:allServos} shows relevant variables for the four cases. For simplicity, we assume there are no constraints in~(\ref{qp}), although off-the-shelf quadratic programming solvers~\cite{nocedal:book:1999} could account for them.

\subsection{Visual Servoing}
\label{Sect:visServo}

\subsubsection{Formulation}
Visual servoing refers to the use of vision to control the robot motion~\cite{chaumette:ram:2006}. The camera may be mounted on a moving part of the robot, or fixed in the workspace. These two configurations are referred to as ``eye-in-hand'' and ``eye-to-hand'' visual servoing, respectively. 
The error $\mathbf{e}$ is defined with regards to some image features, here denoted by $\mathbf{s}$, to be regulated to a desired configuration $\mathbf{s}^*$ ($\mathbf s$ is analogous to $\mathbf x$ in the inverse kinematic formulation above). The visual error is: 
\begin{equation}
\mathbf{e} = \dot{\mathbf{s}} - \dot{\mathbf{s}}^*.
\label{eq:visServo}
\end{equation}

Visual servoing schemes are called \textit{image-based} if $\mathbf{s}$ is defined in image space, and \textit{position-based} if $\mathbf s$ is defined in the 3D operational space. Here we only briefly recall the image-based approach (on its eye-in-hand modality), since the position-based one consists in projecting the task from the image to the operational space to obtain $\mathbf x$  and then apply~(\ref{Eq:taskControl}).

\begin{figure}[t!]
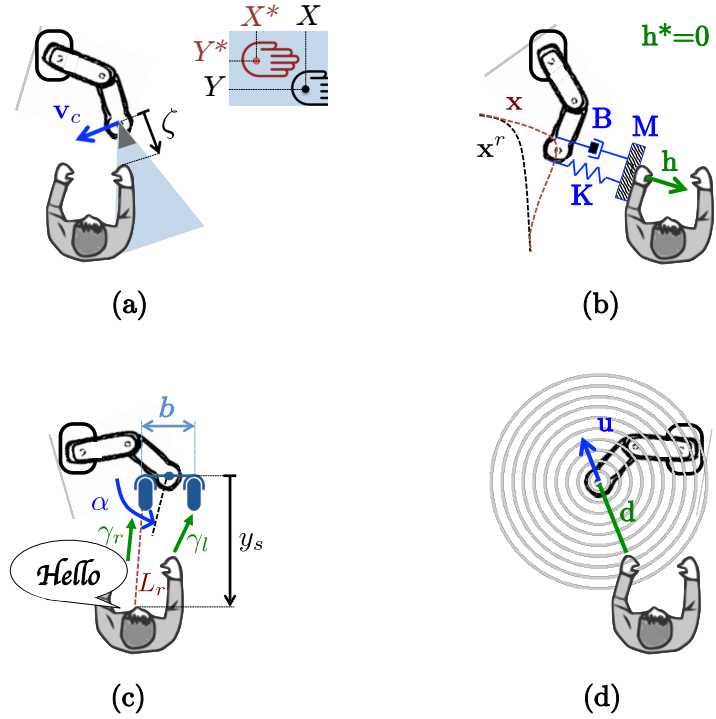

				\centering \FallServos
				\caption{Examples of four sensor-based servo controllers. (a) Visual servoing: the user hand is centered in the camera image. (b) Indirect force control: by applying a wrench, the user deviates the contact point away from a reference trajectory. (c) Audio-based control: a microphone rig is automatically oriented towards the sound source (the user's mouth). (d) Distance-based control: the user acts as a repulsive force, related to his/her distance from the robot.} 
				\label{Fig:allServos}
\end{figure}

The simplest image-based controller uses $\mathbf{s}=[X,Y{]}^\top$, with $X$ and $Y$ as the coordinates of an image pixel, to generate $\mathbf u$ that drives $\mathbf s$ to a reference $\mathbf s^* = [X^*, Y^*]^\top$ (in Fig. \ref{Fig:allServos}a the centroid of the human hand). This is done by defining $\mathbf{e}$ as:
\begin{equation}
\dot{\mathbf s} - \dot{\mathbf s}^* \!=\! 
\left[ \begin{array}{c}
\dot X - \dot{X}^* \\
\dot Y - \dot{Y}^*
\end{array} \right],
\text{with } 
\dot{\mathbf s}^* \!=\!
- \lambda 
\left[ \begin{array}{c}
X - X^* \\
Y - Y^*
\end{array} \right]
\end{equation}
If we use the camera's 6D velocity as the control input $\mathbf{u} = \mathbf{v}_c$, the image Jacobian matrix\footnote{Also known as \textit{interaction matrix} in the visual servoing literature.} relating $[ \dot{X} , \dot{Y}]^\top$ and ${\mathbf{u}}$ is:
\begin{equation}
\mathbf{J}_v = \left[ 
\begin{array}{cccccc}
-\frac{1}{\zeta} 	& 0 			& \frac{X}{\zeta} & XY & -1-X^2 	& Y
\vspace{0.1cm}\\
0 				&-\frac{1}{\zeta} 	& \frac{Y}{\zeta} & 1+Y^2 & -XY 	& -X
\end{array}
\right],
\end{equation}
where $\zeta$ denotes the depth of the point with respect to the camera. In the absence of constraints, the solution of~\eqref{qp} is:
\begin{equation}
\mathbf{v}_c = - \mathbf{J}_v^+ \lambda \left[ \begin{array}{c}
X - X^* \\
Y - Y^* 
\end{array}
\right].
\label{Eq:visControl}
\end{equation}

\subsubsection{Application to Human-Robot Collaboration}

Humans generally use vision to teach the robot relevant configurations for collaborative tasks. For example,~\cite{Cai2016} demonstrates an application where a human operator uses a QR code to specify the target poses for a 6 degrees-of-freedom (dof) robot arm. In \cite{Gridseth2016} the user provides the target tasks via a tablet-like interface that shows the robot the desired reference view. The human can specify various motions such as point-to-point, line-to-line, etc., that are automatically performed via visual feedback. The authors of~\cite{Gridseth2015} present a grasping system for a tele-operated dual arm robot. The user specifies the object to be manipulated, and the robot completes the task using visual servoing.

Assistive robotics represents another very common application domain for visual servoing. The motion of robotic wheelchairs has been semi-automated at various degrees. For instance,~\cite{Narayanan:2016} presents a corridor following method that exploits the projection of parallel lines. The user provides target directions with a haptic interface, and the robot corrects the trajectories with visual feedback.
Other works focus on mobile manipulation. The authors of~\cite{Tsui:2011} develop a vision-based controller for a robotic arm mounted on a wheelchair: the user specifies the object to be grasped and retrieved by the robot. 
A similar approach is reported in \cite{Dune:2008}, where, the desired poses are input with ``clicks'' on an screen interface.

Medical robotics is another area that involves sensor-based interactions between humans and robots, and where vision has huge potential (see~\cite{Azizian2014} for a comprehensive review).
The authors of~\cite{Agustinos2014} design a laparoscopic camera, which regulates its pan/tilt motions to track human-held instruments.

\subsection{Touch (or Force) Control} 

\subsubsection{Formulation}
Touch (or force) control requires the measurement of one or multiple (in the case of tactile skins) \textit{wrenches} $\mathbf{h}$, 
which are (at most) composed of 3 translational forces, and 3 torques; $\mathbf{h}$ is fed to the controller that moves the robot so that it exerts a desired interaction force with the human or environment. 
Force control strategies can be grouped into the following two classes~\cite{villani:hbr:2008}:  
\begin{itemize}
	\item
	\textit{Direct} control regulates the contact wrench to obtain a desired wrench $\mathbf{h}^*$. Specifying $\mathbf{h}^*$ requires an explicit model of the task and environment. A widely adopted strategy is hybrid position/force control~\cite{RaCr:81}, which regulates the velocity and wrench along unconstrained and constrained task directions, respectively. 
	Referring to~\eqref{qp}, this is equivalent to setting 
    \begin{equation}
    \mathbf{e} = \mathbf{S} \left(\dot{\mathbf{x}}-\dot{\mathbf{x}}^*\right) + \left( \mathbf{I} - \mathbf{S} \right) \left(\mathbf h - \mathbf h^*\right),
    \label{eq:dirForceControl}
    \end{equation}
with $\mathbf{S}=\mathbf{S}^\top\ge 0$ a binary diagonal selection matrix, and $\mathbf I$ as the identity matrix. Applying a motion $\mathbf{u}$ that nullifies $\mathbf{e}$ in~(\ref{eq:dirForceControl}) guarantees that the components of $\dot{\mathbf{x}}$ (respectively ${\mathbf{h}}$) specified via $\mathbf{S}$ (respectively $\mathbf{I} - \mathbf{S}$) converge to $\dot{\mathbf{x}}^*$ (respectively ${\mathbf{h}}^*$). 
	\item
\textit{Indirect} control (illustrated in Fig. \ref{Fig:allServos}b) does not require an explicit force feedback loop. To this category belong \textit{impedance control} and its dual \textit{admittance control}~\cite{Ho:85}. It consists in modelling the deviation of the contact point from a reference trajectory $\mathbf{x}^r\left( t \right)$ associated to the desired $\mathbf{h}^*$, via a virtual mechanical impedance with adjustable parameters (inertia $\mathbf{M}$, damping $\mathbf{B}$ and stiffness $\mathbf{K}$). Referring to~\eqref{qp}, this is equivalent to setting: 
					
	\begin{equation}
	\mathbf{e} = \mathbf{M} (\ddot{\mathbf x} - \ddot{\mathbf x}^r)
	+ \mathbf{B} (\dot{\mathbf x} - \dot{\mathbf x}^r)
	+ \mathbf{K} (\mathbf{x} - \mathbf{x}^r) -(\mathbf{h} - \mathbf h^*).
	\label{eq:impedance control}
	\end{equation} 
	
Here, $\mathbf{x}$  represents the ``deviated'' contact point pose, with $\dot{\mathbf x}$ and $\ddot{\mathbf x}$ as time derivatives. When $\mathbf{e} = \mathbf{0}$, the displacement $\mathbf{x} - \mathbf{x}^r$ responds as a mass-spring-damping system under the action of an external force $\mathbf{h} - \mathbf{h}^*$. In most cases, $\mathbf{x}^r\left( t \right)$ is defined for motion in free space ($\mathbf{h}^* = \mathbf{0}$). The general formulation in~\eqref{qp} and~(\ref{eq:impedance control}) can account for both impedance control ($\mathbf{x}$ is measured and $\mathbf{u} = \mathbf{h}$) and admittance control ($\mathbf{h}$ measured and $\mathbf{u} = \mathbf{x}$). 
\end{itemize}

\subsubsection{Application to Human-Robot Collaboration}

The authors of~\cite{Bauzano2016} use direct force control for collaborative human-robot laparoscopic surgery. They control the instruments with a hybrid position/force approach. In~\cite{Cortesao2017}, a robot regulates the applied forces onto a beating human heart. Since the end-effector's 3 linear dof are fully-constrained, position control cannot be performed: $\mathbf{S} = \mathbf{0}$ in~(\ref{eq:dirForceControl}).

A drawback of direct control is that it can realize only the tasks which can be described via constraint surfaces. If their location is unknown and/or the contact geometry is complex---as often in human-robot collaboration---indirect control is more suited since: i) it allows to define a priori how the robot should react to unknown external force disturbances, ii) it can use a reference trajectory $\mathbf{x}^r\left( t \right)$ output by another sensor (e.g., vision). In the next paragraph, we review indirect force control methods.

By sensing force, the robot can infer the motion commands (e.g., pushing, pulling) from the human user. For example, Maeda et al.~\cite{maeda:iros:2001} use force sensing and human motion estimation (based on minimum jerk) within an indirect (admittance) control framework for cooperative manipulation. In~\cite{ErT.se:10,ErMa:11}, an assistant robot suppresses the involuntary vibrations of a human, who controls welding direction and speed. By exploiting kinematic redundancy,~\cite{F.:13} also addresses manually guided robot operation. The papers~\cite{Wang2015,bussy:iros:2012} present admittance controllers for two-arm robots moving a table in collaboration with a human. 
In~\cite{Baumeyer:2015}, a human can control a medical robot arm, with an admittance controller. Robot \textit{Tele-operation} is another common human-robot collaboration application where force feedback plays a crucial role; \cite{Passenberg2010} is a comprehensive review on the topic.

All these works rely on local force/moment measures. To date, tactile sensors and skins (measuring the wrench along the robot body, see \cite{Argall2010} for a review) have been used for object exploration~\cite{NaTo:06} or recognition~\cite{AbChCr:18}, but not for control as expressed in~(\ref{qp}). One reason is that they are at a preliminary design stage, which still requires complex calibration~\cite{CaDeDe:11,LiFiLo:13} that constitutes a research topic per se. An exception is~\cite{HaLiRi:13}, which uses tactile measures within~(\ref{qp}). Similarly, in~\cite{Zhang2000}, tactile sensing regulates interaction with the environment. Yet, neither of these works considers pHRI. In our opinion, there is huge potential in the use of skins and tactile displays for human-robot collaboration. 

\subsection{Audio-Based control}\label{sec:audio-based} 

\subsubsection{Formulation}

The purpose of audio-based control is to locate the sound source, and move the robot towards it. 
For simplicity, we present the two-dimensional binaural (i.e., with two microphones) configuration in Fig. \ref{Fig:allServos}c, with the angular velocity of the microphone rig as control input: $\mathbf u = \dot{\alpha}$. 
We hereby review the two most popular methods for defining error $\mathbf{e}$ in~(\ref{qp}): 
Interaural Time Difference (ITD) and Interaural Level Difference (ILD)\footnote{Or its frequency counterpart: Interaural Phase Difference (IPD).}. 
The following is based on~\cite{Magassouba:2016_wk}:

\begin{itemize}
\item \textit{ITD-based aural servoing} uses the difference $\tau$ between the arrival times of the sound on each microphone; $\tau$ must be  
regulated to a desired $\tau^*$. 
The controller can be represented with~\eqref{qp}, by setting $\mathbf e = \dot{\tau} - \dot{\tau}^*$, with the desired rate $\dot{\tau}^*=-\lambda\left(\tau - \tau^* \right)$ (to obtain set-point regulation to $\tau^*$). 
Feature $\tau$ can be derived in real-time by using standard cross-correlation of the signals~\cite{Youssef2012}. Under a far field assumption:
\begin{equation}
	\mathbf e = \dot{\tau} - \dot{\tau}^*= -\left( \sqrt{(b/c)^2 - \tau^2} \right) \mathbf u - \dot{\tau}^*
	\label{eq:ITD_error}
\end{equation}
with $c$ the sound celerity and $b$ the microphones baseline. From \eqref{eq:ITD_error}, the \textit{scalar} ITD Jacobian is: $\mathbf J_\tau = - \sqrt{(b/c)^2 - \tau^2}$. The motion that minimizes $\mathbf e$ is:
\begin{equation}
	\mathbf u = - \lambda \mathbf J_\tau^{-1}(\tau - \tau^*), 
        \label{eq:audio1}
\end{equation}
which is locally defined for $\alpha\in(0,\pi)$, to ensure that $|\mathbf J_\tau|\ne 0$.

\item \textit{ILD-based aural servoing} uses $\rho$, the difference in intensity between the left and right signals. This can be obtained in a time window of size $N$ as $\rho=E_l/E_r$, where the $E_{l,r}=\sum_{n=0}^N\gamma_{l,r}[n]^2$ denote the signals' sound energies and the $\gamma_{l,r}[n]$ are the intensities at iteration $n$. To regulate $\rho$ to a desired $\rho^*$, one can set $\mathbf e = \dot\rho - \dot\rho^*$ with $\dot{\rho}^*=-\lambda\left(\rho - \rho^* \right)$. Assuming spherical propagation and slowly varying signal:
				
\begin{equation}
	\mathbf e = \dot\rho - \dot\rho^* = \frac{y_s(\rho + 1)b}{L_r^2} \mathbf u - \dot\rho^*
	\label{eq:ILD_error}
\end{equation}
where $y_s$ is the sound source frontal coordinate in the moving auditory frame, and $L_r$ the distance between the right microphone and the source. From \eqref{eq:ILD_error}, the \textit{scalar} ILD Jacobian is $\mathbf J_\rho = y_s(\rho +1)b/L_r^2$. The motion that minimizes $\mathbf e$ is:
\begin{equation}
	\mathbf u = - \lambda \mathbf J_\rho^{-1}(\rho - \rho^*) 
    \label{eq:audio2}
\end{equation}
where $\mathbf J_\rho^{-1}$ is defined for sources located in front of the rig. In contrast with ITD-servoing, here the source location (i.e., $y_s$ and $L_r$) must be known or estimated.
\end{itemize}
While the methods above only control the angular velocity of the rig ($\mathbf u = \dot{\alpha}$), Magassouba has extended both to also regulate the 2D translations of a mobile platform, (ITD in~\cite{Magassouba2015,Magassouba2016} and ILD in~\cite{Magassouba:2016}).

\subsubsection{Application to Human-Robot Collaboration}

Due to the nature of this sense, audio-based controllers are mostly used in contact-less applications, 
to enrich other senses (e.g., force, distance) with sound, or to design intuitive interfaces. 

Audio-based control is currently (in our opinion) an underdeveloped research area with great potential for human-robot collaboration, e.g., for tracking a speaker.
Besides the cited works \cite{Magassouba:2016_wk,Magassouba2015,Magassouba2016,Magassouba:2016}, that closely follow the framework of Sec. \ref{chap:sensor-based_control}, others formulate the problem differently. 
For example, the authors of \cite{Kumon2003,Kumon2005} 
propose a linear model to describe the relation between the 
pan motion of a robot head and the difference of intensity between its two microphones. 
The resulting controllers are much simpler 
than~(\ref{eq:audio1}) and~(\ref{eq:audio2}). Yet, their operating range is smaller, making them less robust than their -- more analytical -- counterparts.



\subsection{Distance-Based control} 
\label{sec:distControl}

\subsubsection{Formulation}

The simplest (and most popular) distance-based controller is the artificial potential fields method~\cite{Kh:85}. Despite being prone to local minima, it has been thoroughly deployed both on manipulators and on autonomous vehicles for obstacle avoidance. 
Besides, 
it is acceptable that a collaborative robot stops (e.g., because of local minima) as long as it avoids the human user. The potential fields method consists in modeling each obstacle as a source of repulsive forces, related to the robot distance from the obstacle (see Fig.~\ref{Fig:allServos}d). All the forces are summed up resulting in a velocity 
in the most promising direction. Given $\mathbf d$, the position of the nearest obstacle in the robot frame, the original version~\cite{Kh:85} consists in applying operational space velocity
\begin{equation}
\mathbf{u} = 
\left\{ 
\begin{array}{ll}
\lambda \left( \frac{1}{\|\mathbf{d}\|} - \frac{1}{d_o} \right) \frac{\mathbf{d}}{\| \mathbf{d} \|^2} & \text{ if } \|\mathbf{d}\| < d_o,\\
0  & \text{ otherwise.}
\end{array}
\right.
\label{eq:potFields}
\end{equation}
Here $d_o > 0$ is the (arbitrarily tuned) minimal distance required for activating the controller.
Since the quadratic denominator in~(\ref{eq:potFields}) yields abrupt accelerations, more recent versions adopt a linear behavior. 
Referring to~\eqref{qp}, this can be obtained by setting $\mathbf{e} = \dot{\mathbf{x}} - \dot{\mathbf{x}}^*$ with $\dot{\mathbf{x}}^* = \lambda \left( 1 - {d}_0 / \|\mathbf{d}\| \right) \mathbf{d}$ as reference velocity:
\begin{equation}
\mathbf{e} = \dot{\mathbf{x}} - \lambda \left( 1 - \frac{{d}_0}{\|\mathbf{d}\|} \right) \mathbf{d}.
\end{equation}
By defining as control input $\mathbf u = \dot{\mathbf{x}}$, the solution to~(\ref{qp}) is:
\begin{equation}
{\mathbf{u}} = \lambda \left( 1 - \frac{{d}_0}{\|\mathbf{d}\|} \right) \mathbf{d}.
\label{eq:repulLin}
\end{equation}

\subsubsection{Application to Human-Robot Collaboration}

Many works use this (or similar) distance-based methods for pHRI. To avoid human-robot collisions, the authors of~\cite{SaLiSi:07} apply~(\ref{eq:repulLin}), by estimating $\mathbf{d}$ between human head and robot with vision. Recently, these approaches have been boosted by the advent of 3D vision sensors (e.g. the Microsoft Kinect and Intel RealSense), which 
can provide both vision and distance control. The authors of~\cite{DeFlKh:12} design a Kinect-based distance controller (again, for human collision avoidance) with an expression similar to~(\ref{eq:repulLin}), but smoothed by a sigmoid.

\textit{Proximity servoing} is a similar technique, which regulates---via capacitive sensors---the distance between the robot surface and the human.  
In~\cite{Schlegl2013}, these sensors 
modify the position and velocity of a robot arm when a human approaches it, to avoid collisions. The authors of~\cite{Bergner2017,Leboutet2016,Dean2017} have developed a new capacitive skin for a dual-arm robot. They design a collision avoidance method based on an admittance model similar to~(\ref{eq:impedance control}), which relies on the joint torques (measured by the skin) to control the robot motion. 

\section{Integration of Multiple Sensors}
\label{sect:integration}

In Sect.~\ref{chap:sensor-based_control}, we presented the most common sensor-based methods used for collaborative robots. Just like natural senses, artificial senses provide complementary information about the environment. Hence, to effectively perform a task, the robot should measure (and use for control) multiple feedback modalities. In this section, we review various methods for integrating multiple sensors in a unique controller.

Inspired by how humans merge their percepts~\cite{ernst:nature:2002}, researchers have traditionally fused heterogeneous sensors to estimate the state of the environment. This can be done in the sensors' Cartesian frames~\cite{smits:mfi:2008} by relying on an Extended Kalman Filter (EKF)~\cite{taylor:book:2006}. 
Yet the sensors must be related to a single quantity, which is seldom the case when measuring different physical phenomena~\cite{nelson:itroa:1996}. An alternative is to use the sensed feedback directly in~(\ref{qp}). This idea, proposed for position-force control in~\cite{RaCr:81} and extended to vision in~\cite{nelson:acc:1995}, brings new challenges to the control design, e.g., sensor synchronization, task compatibility and task representation. For instance, the designer should take care when transforming $6$ D velocities or wrenches to a unique frame. This requires (when mapping from frame $A$ to frame $B$) multiplication by
\begin{equation}
^B\mathbf V_A
= \begin{bmatrix}
^B\mathbf{R}_A & \left[ ^B\mathbf{t}_A \right]_\times ^B\mathbf{R}_A  \\
\mathbf{0}_3 & ^B\mathbf{R}_A
\end{bmatrix}
\label{eq:transformV}
\end{equation}
for a velocity, and by $^B\mathbf V^\top_A$ 
for a wrench.
In~(\ref{eq:transformV}), $^B\mathbf{R}_A$ is the rotation matrix from $A$ to $B$ and $\left[ ^B\mathbf{t}_A \right]_\times$ the skew-symmetric matrix associated to translation $^B\mathbf{t}_A$.

\begin{figure}[t!]
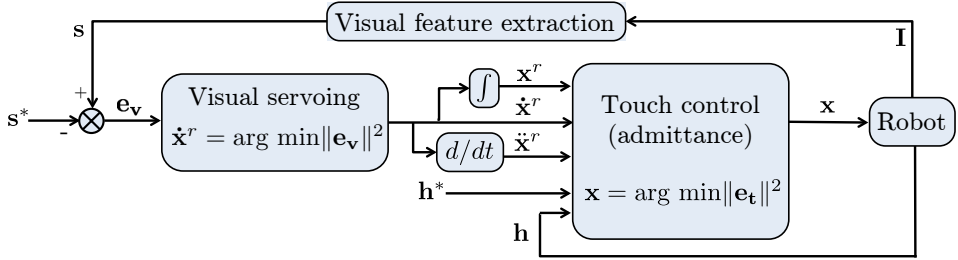

    	\centering \FsharedVisionForce
    	\caption{The most common scheme for \textit{shared} vision/touch (admittance) control, used in~\cite{morel:icra:1998}, \cite{c22,c25}. The goal is to obtain desired visual features ${\mathbf{s}}^*$ and wrench ${\mathbf{h}}^*$, based on current image ${\mathbf{I}}$ and wrench ${\mathbf{h}}$. The \textit{outer} visual servoing loop based on error~(\ref{eq:visServo}) outputs a reference velocity $\dot{\mathbf{x}}^r$ that is then deformed by the \textit{inner} admittance control loop based on error~(\ref{eq:impedance control}), to obtain the desired robot position ${\mathbf{x}}$.} 
    	\label{Fig:FsharedVisionForce}
    \end{figure}

According to~\cite{nelson:acc:1995}, the three methods for combining $N$ sensors within a controller are: 
\begin{itemize}
	\item 
	\textit{Traded}: the sensors control the robot one at a time. Predefined conditions on the task trigger the switches:
		     \begin{equation}
		     \mathbf{u} = \left\{\begin{array}{ll}
		     \underset{\mathbf{u}}{\argmin}~ \| \mathbf{e}_1( \mathbf{u} ) \|^2 & 
				 {\text{if } (\textit{condition 1})=\text{true,}}\\
		     \vdots \\
		     \underset{\mathbf{u}}{\argmin}~ \| \mathbf{e}_N ( \mathbf{u} ) \|^2 &
				 {\text{if } (\textit{condition N})=\text{true.}}
		     \end{array}
		     \right.
		     \end{equation}
  
    \item
	\textit{Shared}: 
	All sensors control the robot 
	throughout operation. A common way is via nested control loops, as shown---for shared vision/touch control---in Fig.~\ref{Fig:FsharedVisionForce}. Researchers have used at most two loops, denoted $o$ for outer and $i$ for inner loop:
    \begin{align}
    \mathbf{u} &= \underset{\mathbf{u}}
    {\argmin}~ \| \mathbf{e}_i \left( \mathbf{u}, \mathbf{u}_o \right) \|^2\\ 
    & \text{ such that }\mathbf{u}_o = \underset{\mathbf{u}_o}
    {\argmin}~ \| \mathbf{e}_o \left( \mathbf{u}_o \right) \|^2 \nonumber.
    \end{align}
In the example of Fig.~\ref{Fig:FsharedVisionForce}: ${\mathbf{u}}={\mathbf{x}}$, ${\mathbf{u}_o}=\dot{\mathbf{x}}^r$, ${\mathbf{e}}_{o} = {\mathbf{e}}_{\tt{\mathbf{v}}}$ applying~(\ref{eq:visServo}) and ${\mathbf{e}}_{i} = {\mathbf{e}}_{\tt{\mathbf{t}}}$ applying~(\ref{eq:impedance control}).

    \item
	\textit{Hybrid}: the sensors act simultaneously, but on different axes of a predefined Cartesian
	\textit{task-frame}~\cite{baeten:ijrr:2003}. The directions are selected by binary diagonal matrices $\mathbf{S}_j$, $j = 1,\dots,N$ with the dimension of the task space, and such that $\sum_{j=1}^{N} \mathbf{S} = \mathbf{I}$:
    \begin{equation}
    \mathbf{u} = \underset{\mathbf{u}}{\argmin}~ \| \sum_{j=1}^{N} \mathbf{S}_j \mathbf{e}_j \left( \mathbf{u} \right) \|^2.
    \label{eq:hybrid}
    \end{equation}
		To express all $\mathbf{e}_j$ in the same task frame, one should apply $^B\mathbf V_A$ and/or $^B\mathbf V^\top_A$. Note the analogy between~(\ref{eq:hybrid}) and the hybrid position/force control framework~(\ref{eq:dirForceControl}).

\end{itemize}
We will use this classification to characterize the works reviewed in the rest of this Section.

\subsection{Traded Control}
\label{tradedControl}

The paper~\cite{j10} presents a human-robot manufacturing cell for collaborative assembly of car joints. The approach (traded \textit{vision/touch}) can manage physical contact between robot and human, and between robot and environment, via admittance control~(\ref{eq:impedance control}). Vision takes over in dangerous situations to trigger emergency stops. The switching \textit{condition} is determined by the position of the human wrt the robot.

In~\cite{Okuno2001,Okuno2004}, a traded \textit{vision/audio} controller enables a mobile robot to exploit sound source localization for visual control. The robot head automatically rotates towards the estimated direction of the human speaker, and then visually tracks him/her. The switching \textit{condition} is that the sound source is visible.
The audio-based task is equivalent to regulating $\tau$ to $0$ or $\rho$ to $1$, as discussed in Sect. \ref{sec:audio-based}. Paper~\cite{Hornstein2006} presents another traded \textit{vision/audio} controller for the iCub robot head to localize a human speaker. This method constructs audio-motor maps based and integrates visual feedback to update the map. Again, the switching \textit{condition} is that the speaker's face is visible. In~\cite{Chan2012},
another traded \textit{vision/audio} controller is deployed on a mobile robot, to drive it towards an unknown sound source; the switching \textit{condition} is defined by a threshold on the frontal localization error.

The authors of \cite{Papageorgiou2014} present a mobile assistant for people with walking impairments. The robot is equipped with: two wrench sensors to measure physical interaction with the human, an array of microphones for audio commands, laser sensors for detecting obstacles, and an RGB-D camera for estimating the users' state. Its controller integrates \textit{audio}, \textit{touch}, \textit{vision} and \textit{distance} 
in a traded manner, with switching \emph{conditions} determined by a knowledge-based layer.

The work \cite{Escaida2014} presents an object manipulation strategy, integrating \textit{distance} (capacitive proximity sensors) and \textit{touch} (tactile sensors). While the method does not explicitly consider humans, it may be applied for human-robot collaboration, since proximity sensors can detect humans if vision is occluded. 
The switching \emph{condition} between the two modes is the contact with the object.

Another example of traded control---here, \textit{audio/distance}---is \cite{Huang1999}, which presents a method for driving a mobile robot towards hidden sound sources, via an omnidirectional array of microphones. The controller switches to ultrasound-based obstacle avoidance in the presence of humans/objects. The detection of a nearby obstacle is the switching \textit{condition}.

\subsection{Shared Control}
\label{sharedControl}

In applications where the robot and environment/human are in permanent contact (e.g. collaborative object transportation), shared control is preferable. Let us first review a pioneer controller~\cite{morel:icra:1998} that relies on shared \textit{vision/touch}, as outlined in Fig.~\ref{Fig:FsharedVisionForce};~\cite{morel:icra:1998} addresses tele-operated peg-in-hole assembly, by placing the visual loop outside the force loop. The reference trajectory $\dot{\mathbf{x}}^r$ output by visual servoing is deformed in the presence of contact by the admittance controller, to obtain the robot position command $\mathbf{x}$. Human interaction is not considered in this work. 

The authors of \cite{Natale2002} estimate sensory-motor responses to control a pan-tilt robot head with shared \textit{visual/audio} feedback from humans. They assume local linear relations between the robot motions and the ITD/ILD measures. This results in controller which is simpler than the one presented in Sect.~\ref{sec:audio-based}. 
The scheme is similar to Fig. \ref{Fig:FsharedVisionForce}, with an outer \textit{vision} loop generating a reference motion, and \textit{audio} modifying it.

\subsection{Hybrid Control}
\label{Sect:hybridControl}

Pomares et al. \cite{Pomares2011} propose a hybrid \emph{vision/touch} controller for grasping objects, using a robot arm equipped with a hand. Visual feedback drives an active camera (installed on the robot tip) to observe the object and detect humans to be avoided, 
whereas touch feedback moves the fingers, to grasp the object. The authors define matrix $\mathbf S$ in~\eqref{eq:dirForceControl} to independently control arm and fingers with the respective sensor.

In~\cite{Chatelain2017}, a hybrid scheme controls an ultrasonic probe in contact with the abdomen of a patient. The goal is to centre the lesions in the ultrasound image observed by the surgeon. The probe is moved by projecting, via $\mathbf{S}$, the \textit{touch} and \textit{vision} (from the ultrasound image) tasks in orthogonal directions.

\subsection{Other Control Schemes}
\label{Sect:combined}

Some works do not strictly follow the classification given above. 
These are reviewed below.

The authors of~\cite{c22,c25} combine \textit{vision} and \textit{touch} to address joint human-humanoid table carrying. The table must stay flat, to prevent objects on top from falling off. Vision controls the table inclination, whereas the forces exchanged with the human make the robot follow his/her intention. The approach is \textit{shared}, with visual servoing in the outer loop of admittance control (Fig.~\ref{Fig:FsharedVisionForce}), to make all dof compliant. However, it is also \textit{hybrid}, since some dof are controlled only with admittance. Specifically vision regulates only the table height in~\cite{c22}, and both table height and roll angle in~\cite{c25}. 

The works~\cite{j6,j7} merge \textit{vision} and \textit{distance} to guarantee lidar-based obstacle avoidance during camera-based navigation. While following a pre-taught path, the robot must avoid obstacles which were not present before. Meanwhile, it moves the camera pan angle, to maintain scene visibility. Here, the selection matrix in~(\ref{eq:hybrid}) is a scalar function $\mathbf{S} \in \left[ 0 , 1 \right]$ dependent on the time-to-collision. In the safe context ($\mathbf{S}=0$), the robot follows the taught path, with camera looking forward. In the unsafe context ($\mathbf{S} = 1$) the robot circumnavigates the obstacles. Therefore, the scheme is \textit{hybrid} when $\mathbf{S}=0$ or $\mathbf{S}=1$ (i.e., vision and distance operate on independent components of the task vector), and \textit{shared} when $\mathbf{S} \in \left(0, 1\right)$.

In~\cite{Dean2016}, proximity (\textit{distance}) and tactile (\textit{touch}) measurements control a robot arm in a pHRI scenario to avoid obstacles or -- when contact is inevitable -- to generate compliant behaviors. The framework linearly combines the two senses, and provides this signal to an inner admittance-like control loop (as in the \emph{shared} scheme of Fig. \ref{Fig:FsharedVisionForce}). Since the operation principle of both senses is complementary (one requires contact while the other does not), the integration can also be seen as \emph{traded}.

The authors of~\cite{j8} enables a robot to adapt to changes in the human behaviour, during a human-robot collaborative screwing task. In contrast with classic hybrid vision--touch--position control, their scheme enables smooth transitions, via weighted combinations of the tasks. The robot can execute \textit{vision} and \textit{force} tasks, either exclusively on different dof (\textit{hybrid} approach) or simultaneously (\textit{shared} approach).

\section{Classification of Works and Discussion}

\begin{table}
\small
\caption{Classification of all papers according to four criteria.} 
\label{tableAll}
\centering
\begin{tabular}
{|c|c|c|c|c|}	
\hline
\textbf{Paper} & \textbf{Sense(s)} &	\textbf{Control objective} & \textbf{Sector} & \textbf{Robot}\\  \hline
\cite{Cai2016}\cite{Gridseth2016} & Vision & Contactless guidance 
& Service & Arm \\ \hline
\cite{Gridseth2015} & Vision & Remote guidance 
& Service & Arm \\ \hline
\cite{Narayanan:2016}-\cite{Dune:2008} & Vision & Contactless guidance 
& Medical & Wheeled \\ \hline
\cite{Agustinos2014} & Vision & Contact w/humans 
& Medical & Arm \\ \hline
\cite{Bauzano2016} & Touch & Contact w/humans 
& Medical & Arm \\
& & Remote guidance 
& & \\ \hline
\cite{Cortesao2017} & Touch & Contact w/humans 
& Medical & Arm \\ \hline
\cite{maeda:iros:2001}-\cite{F.:13} & Touch & Direct guidance 
& Production & Arm \\ \hline
\cite{Wang2015} & Touch & Carrying 
& Production & Wheeled \\ \hline
\cite{bussy:iros:2012} & Touch & Carrying 
& Production & Humanoid \\ \hline
\cite{Baumeyer:2015} & Touch & Remote guidance 
& Medical & Arm \\ \hline
\cite{Magassouba:2016_wk}\cite{Kumon2003}\cite{Kumon2005} & Audition & Contactless guidance 
& Service & Heads \\ \hline
\cite{Magassouba2015}-\cite{Magassouba:2016} & Audition & Contactless guidance 
& Service & Wheeled \\ \hline
\cite{SaLiSi:07}-\cite{Schlegl2013} & Distance & Collision avoidance 
& Production & Arm \\ \hline
\cite{Bergner2017}-\cite{Dean2017} & Distance & Collision avoidance 
& Service & Arm \\ \hline
\cite{j10} & V+T~(tra.) & Assembly 
& Production & Arm \\ \hline
\cite{Okuno2001}-\cite{Hornstein2006} & V+A(tra.) & Contactless guidance 
& Service & Heads \\ \hline
\cite{Chan2012} & V+A(tra.) & Contactless guidance 
& Service & Wheeled \\ \hline
\cite{Papageorgiou2014} & V+T+A+D & Direct guidance 
& Medical & Wheeled \\
& (tra.)&&& \\ \hline
\cite{Escaida2014} & D+T(tra.) & Collision avoidance 
& Production & Arm \\  \hline
\cite{Huang1999} & D+A(tra.) & Collision avoidance 
& Service & Wheeled \\ \hline
\cite{Natale2002} & V+A(sh.) & Contactless guidance 
& Service & Heads \\ \hline
\cite{Pomares2011} & V+T(hyb.) & Collision avoidance 
& Production & Arm \\ \hline
\cite{Chatelain2017} & V+T & Contact w/humans 
& Medical & Arm \\
&(hyb.)&Remote guidance 
&&\\ \hline
\cite{c22}\cite{c25} & V+T & Contact w/humans 
& Production & Humanoid \\
&(sh.+hyb.)&&&\\ \hline
\cite{j6}\cite{j7} & D+V & Collision avoidance 
& Production & Wheeled \\
&(sh.+hyb.)&&&\\ \hline
\cite{Dean2016} & D+T & Direct guidance 
& Service & Arm \\
&(sh.+tra.)&&&\\ \hline
\cite{j8} & V+T & Assembly 
& Production & Arm \\
&(sh.+hyb.)&&&\\ \hline
\end{tabular}
\end{table}

In this section, we use five criteria to classify all the surveyed papers which apply sensor-based control to collaborative robots. This taxonomy then serves as an inspiration to drive the following discussion on design choices, limitations and future challenges. 

In total, we refer to the forty-five papers revised above. These include the works with only one sensor, discussed in Sect.~\ref{chap:sensor-based_control} (\cite{Cai2016}--\cite{Baumeyer:2015}, \cite{Magassouba:2016_wk}--\cite{Dean2017}) and those which integrate multiple sensors, discussed in Sect.~\ref{sect:integration} (\cite{j10}--\cite{j8}). The five criteria are: sensor(s), integration method (when multiple sensors are used), control objective, target sector and robot platform. In Table~\ref{tableAll}, we indicate these characteristics for each paper. Then, we focus on each characteristic, in Tables~\ref{tableSoA}-\ref{tablePlatform}\footnote{In the Tables, we have used the following notation: V, T, A, D for Vision, Touch, Audition and Distance, and sh., hyb., tra. for shared, hybrid and traded.}. 

Table \ref{tableSoA} classifies the papers according to the sensor/s. Column \textit{mono} indicates the papers relying only on one sensor. For the others, we specify the integration approach (see Sect.~\ref{sect:integration}). Note that vision (alone or not) is by far the most popular sense, used in 22 papers. 
This comes as no surprise, since even for humans, vision provides the richest perceptual information to structure the world and perform motion \cite{hoffman1998visual}.		
Touch is the second most commonly used sensor (18 papers) and fundamental in pHRI, since it is the only one among the four that can be exploited directly to modulate contact. 

Also note that, apart from~\cite{Papageorgiou2014}, no paper integrates more than two sensors. Given the sensors wide accessibility and the recent progress in computation power, this is probably due to the difficulty in designing a framework capable of managing such diverse and broad data. 
Another reason may be the presumed (but disputable) redundancy of the three contact-less senses, which biases towards opting for vision, given its diffusion and popularity (also in terms of software). Touch -- the only sensor measuring contact -- is irreplaceable. This may also be the reason why, when merging two sensors, researchers have generally opted for vision+touch (7 out of 17 papers). 
The most popular among the three integration methods is \textit{traded control}, probably because it is the easiest to set up. In recent years, however, there has been a growing interest towards the \textit{shared+hybrid} combination, which guarantees nice properties in terms of control smoothness.

\begin{table}[t!]
	\caption{Classification based on the sensors.}
	\label{tableSoA}
	\centering
	\small
	\begin{tabular}%
	{|*{5}{c|}}
		\cline{1-2}
		Vision 				& 
		\cite{Cai2016}-\cite{Agustinos2014}
		\\ \cline{1-3}
		Touch 			
		& 
		\begin{tabular}{@{}c@{}}
			\cite{Bauzano2016}-\cite{Baumeyer:2015}
		\end{tabular}
		& 
		\begin{tabular}{@{}c@{}}
			tra.~\cite{j10} hyb.~\cite{Pomares2011}\cite{Chatelain2017} \\ 
			sh.+hyb.~\cite{c22}\cite{c25}\cite{j8}       \end{tabular}
		\\ \cline{1-4}
		Audition 				
		&   \cite{Magassouba:2016_wk}-\cite{Kumon2005}
		& 
		\begin{tabular}{@{}c@{}}
			tra. \cite{Okuno2001}-\cite{Papageorgiou2014} sh. \cite{Natale2002}
		\end{tabular}
		& 
		tra. \cite{Papageorgiou2014} \\
		\cline{1-5}
		Distance 			
		& 
		\begin{tabular}{@{}c@{}}
			\cite{SaLiSi:07}-\cite{Dean2017}
		\end{tabular}
		& 
		\begin{tabular}{@{}c@{}}
			sh.+hyb.~\cite{j6}\cite{j7}
		\end{tabular}
		& sh.+tra.~\cite{Dean2016}
		& tra.~\cite{Papageorgiou2014}\cite{Huang1999}     \\
		&&& tra. \cite{Escaida2014}&\\ \hline
		& Mono & Vision & Touch & Audition \\ \hline
	\end{tabular}
\end{table}

An unexploited application of shared control is the combination of \textit{vision} and \textit{distance} (proximity sensors) to avoid collisions with humans. This can be formulated as in Fig.~\ref{Fig:FsharedVisionForce} by replacing touch control error $\mathbf{e_t}$ with an admittance-like distance control error:
\begin{equation}
\mathbf{e}_d =  -( \mathbf d - \mathbf d^*) + \mathbf{M} (\ddot{\mathbf x} - \dot{\mathbf x}^r)
+ \mathbf{B} (\dot{\mathbf x} - \dot{\mathbf x}^r)
+ \mathbf{K} (\mathbf{x} - \mathbf{x}^r),
\label{eq:proxError}
\end{equation}
where $\mathbf d$ and $\mathbf d^*$ represent the measured and desired distance to obstacles. With this approach, the robot can stabilize at a given ``safe'' distance from an obstacle, or move away from it.

In the authors' opinion, no sensor(s) nor (if needed) integration method is the best, and the designer should choose according to the objective at stake. For this, nature and evolution can be extremely inspiring but technological constraints (e.g., hardware and software availability) must also be accounted for, with the golden rule of engineering that ``simpler is better''.

Table \ref{tableObj} classifies the papers according to the \textit{control objective}. In the table, we also apply the taxonomy of pHRI layers introduced in~\cite{LuFl:12}, and evoked in the introduction: \textit{safety}, \textit{coexistence}, \textit{collaboration}.
Works that focus on collision avoidance address \textit{safety}, and works where the robot acts on passive humans address \textit{coexistence}. For the \textit{collaboration} layer, we distinguish two main classes of works. First, those where the human is guiding the robot (without contact, with direct contact, or with remote physical contact as in tele-operation), then those where the two collaborate 
(e.g., for part assembly or object carrying). The idea (also in line with~\cite{LuFl:12}) is the lower lines in the table generally require higher cognitive capabilities (e.g., better modelling of environment and task). Some works, particularly in the field of medical robotics~\cite{Agustinos2014,Bauzano2016,Chatelain2017} cover both coexistence and collaboration, since the human is guiding the robot to operate on another human. 
Interestingly, the senses appear in the table with a trend analogous to biology. \textit{Distance} is fundamental for collision avoidance, when the human is far, and his/her role in the interaction is basic (s/he is mainly perceived as an obstacle). Then, audio is used for contactless guidance. As human and robot are closer, \textit{touch} takes over the role of \textit{audio}. 
As mentioned above, \textit{vision} is a transversal sense, capable of covering most distance ranges. Yet, when contact is present (i.e., in the four lower lines), it is systematically complemented by touch, a popular pairing as also shown in Table~\ref{tableSoA} and discussed above. 

\begin{table}
	\caption{Classification based on the control objective with corresponding pHRI layer as proposed in \cite{LuFl:12} (in parenthesis).} 
	\label{tableObj}
	\centering
	\small
	\begin{tabular}{|c|c|}
		\cline{1-2}
		Collision avoidance  &  distance \cite{SaLiSi:07}-\cite{Dean2017} distance+touch \cite{Escaida2014} \\ 
		(\textit{safety}) & distance+audition \cite{Huang1999} vision+touch \cite{Pomares2011}\\
		&  vision+distance~\cite{j6,j7} \\ 
		\hline
		Contact with passive humans & vision~\cite{Agustinos2014} touch \cite{Bauzano2016,Cortesao2017}\\ 
		(\textit{coexistence}) & vision+touch~\cite{Chatelain2017}\\ 
		\hline
		Contactless guidance & vision \cite{Cai2016,Gridseth2016}, \cite{Narayanan:2016}-\cite{Dune:2008}\\ 
		(\textit{collaboration}) & audition \cite{Magassouba:2016_wk}-\cite{Kumon2005}\\ 
		& vision+audition \cite{Okuno2001}-\cite{Chan2012},~\cite{Natale2002} \\
		\hline
		Direct guidance &  touch+audition+distance+vision \cite{Papageorgiou2014} \\ 
		(\textit{collaboration}) & touch~\cite{maeda:iros:2001}-\cite{F.:13} touch+distance \cite{Dean2016} \\ 
		\hline
		Remote guidance & vision~\cite{Gridseth2015,Agustinos2014} touch~\cite{Bauzano2016,Baumeyer:2015}\\ 
		(\textit{collaboration}) &vision+touch~\cite{Chatelain2017}\\ 
		\hline
		Collaborative assembly & vision+touch~\cite{j10,j8} \\ 
		(\textit{collaboration}) & \\ 
		\hline
		Collaborative carrying & touch~\cite{Wang2015,bussy:iros:2012} \\ 
		(\textit{collaboration}) & vision+touch~\cite{c22, c25} \\ 
		\hline
	\end{tabular}
\end{table}

Table \ref{tableApp} classifies the papers according to the target (or \emph{potential}) sector. 
We propose three sectors: \emph{Production}, \emph{Medical}, and \emph{Service}. 
Production is the historical sector of robotics; applications include: manufacturing (assembly, welding, pick-and-place), transportation (autonomous guided vehicles, logistics) and construction (material and brick transfer).
The medical category has become very popular in recent years, with applications spanning from robotic surgery (surgical gripper and needle manipulation), diagnosis (positioning of ultrasonic probes or endoscopes), and assistance (intelligent wheelchairs, feeding and walking aids).
The service sector is the one that in the authors' opinion presents the highest potential for growth in the coming years.
Applications include companionship (elderly and child care), domestic (cleaning, object retrieving), personal (chat partners, tele-presence). 
The table shows that all four sensors have been deployed in all three sectors. The only exception is \textit{audition} not being used in \textit{production} applications, probably because of the noise -- common in industrial environments.

Finally, Table \ref{tablePlatform} gives a classification based on the robotic platform.
We can see that (unsurprisingly) most works use fixed base \emph{arms}. 
The second most used platforms here are \emph{wheeled} robots. Then, the \emph{humanoids} category, which refers to robots with anthropomorphic design (two arms and biped locomotion capabilities). Finally, we consider robot \emph{heads}, which are used exclusively for audio-based control. While robot heads are commonly used for face tracking in \textit{Social Human Robot Interaction}, such works are not reviewed in this survey as they do not generally involve contact.


\begin{table}[t]
	\caption{Classification based on target/potential sectors.}
	\label{tableApp}
	\centering
	\small
	\begin{tabular}{|*{2}{c|}}
		\cline{1-2}
		\emph{Production} (manufacturing, &  touch~\cite{maeda:iros:2001}-\cite{bussy:iros:2012} distance \cite{SaLiSi:07}-\cite{Schlegl2013} \\
		transportation, construction) & D+T \cite{Escaida2014} V+T~\cite{j10}\cite{Pomares2011}\cite{c22}\cite{c25}\cite{j8}\\
		& V+D~\cite{j6,j7}\\
		\hline
		\emph{Medical} (surgery, diagnosis, & vision~\cite{Narayanan:2016}-\cite{Agustinos2014} touch~\cite{Bauzano2016}\cite{Cortesao2017}\cite{Baumeyer:2015}\\  
		assistance)& V+T+A+D \cite{Papageorgiou2014} V+T~\cite{Chatelain2017}\\
		\hline
		\emph{Service} (companionship, & vision \cite{Cai2016}-\cite{Gridseth2015} audition \cite{Magassouba:2016_wk}-\cite{Kumon2005}  \\
		domestic, personal) & distance \cite{Bergner2017}-\cite{Dean2017} V+A \cite{Okuno2001}-\cite{Chan2012},~\cite{Natale2002} \\
		&D+A \cite{Huang1999} T+D \cite{Dean2016} \\
		\hline
	\end{tabular}
\end{table}

\section{Conclusions}


This work presents a systematic review of sensor-based controllers which enable collaboration and/or interaction between humans and robots.
We considered four senses: vision, touch, audition and distance.
First, we introduce a tutorial-like general formulation of sensor-based control, which we instantiate for visual servoing, touch control, aural servoing, and distance-based control, while reviewing representative papers.
Next, with the same formulation, we model the methods that integrate multiple sensors, while again discussing related works.
Finally, we classify the surveyed body of literature according to: used sense(s), integration method, control objective, target application and platform.

Althoug vision and touch (\textit{proprioceptive force} rather than \textit{tact}) emerge nowadays as the most popular senses on collaborative robots, the advent of cheap, precise and easy to integrate tactile, distance and audio sensors present great opportunities for the future.
Typically, we believe that robot skins (e.g., on arms and hands) will simplify interaction, boosting the opportunities for human-robot collaboration. 
It is imperative that researchers develop the appropriate tools for this. Distance/proximity feedback is promising to fully perceive the human operating near the robot (something monocular vision cannot do). Audio feedback is key for developing robotic heads that can interact in a natural way with human speakers. 

\begin{table}[t]
	\caption{Classification based on the type of robot platform.}
	\label{tablePlatform}
	\centering
	\small
	\begin{tabular}
		{|*{2}{c|}}
		\cline{1-2}
		Arms & vision~\cite{Cai2016}-\cite{Gridseth2015},~\cite{Agustinos2014}  touch~\cite{Bauzano2016}-\cite{F.:13},~\cite{Baumeyer:2015} distance~\cite{SaLiSi:07}-\cite{Dean2017} \\
		&  V+T~\cite{j10,Pomares2011,Chatelain2017,j8} D+T \cite{Escaida2014,Dean2016} \\
		\hline
		& vision~\cite{Narayanan:2016}-\cite{Dune:2008} touch~\cite{Wang2015} audition~\cite{Magassouba2015}-\cite{Magassouba:2016} \\
		Wheeled & V+A~\cite{Chan2012} V+T+A+D \cite{Papageorgiou2014} D+A~\cite{Huang1999} V+D~\cite{j6,j7} \\
		\hline
		Humanoids & touch~\cite{bussy:iros:2012} V+T~\cite{c22,c25} \\
		\hline
		Heads & audition~\cite{Magassouba:2016_wk,Kumon2003,Kumon2005} V+A~\cite{Okuno2001}-~\cite{Hornstein2006}~\cite{Natale2002}  \\
		\hline
	\end{tabular}
\end{table}

Finally, some open problems must be addressed, to develop robust controllers for real-world applications. For example, the use of task constraints has not been sufficiently explored when multiple sensors are integrated.
Also, difficulty in obtaining models describing and predicting human behavior hampers the implementation of human-robot collaborative tasks. 
The use of multimodal data such as RGB-D cameras with multiple proximity sensors may be an interesting solution for this human motion sensing and estimation problem. More research needs to be conducted in this direction.


\bibliographystyle{IEEEtran}
\bibliography{library}

\end{document}